%% file: acl2023.tex
\pdfoutput=1
\documentclass[11pt]{article}

\usepackage{ACL2023}
\usepackage{booktabs}
\usepackage{times}
\usepackage{latexsym}
\usepackage{stfloats}

\usepackage[T1]{fontenc}
\usepackage[utf8]{inputenc}

\usepackage{microtype}

\usepackage{inconsolata}

\usepackage{fancyhdr}
\usepackage[T1]{fontenc}
\usepackage[normalem]{ulem}
\usepackage{adjustbox}
\usepackage{booktabs}
\usepackage{multirow}

\usepackage[labelfont=bf]{caption}
\useunder{\uline}{\ul}{}

\title{\textsc{FinanceBench}: A New Benchmark for Financial Question Answering}
\author{Pranab Islam$^{1}$\thanks{\;\;Authors are ordered alphabetically} \quad Anand Kannappan$^{1}$\quad Douwe Kiela$^{2,3}$ \\ \textbf{Rebecca Qian}$^{1}$\quad \textbf{Nino Scherrer}$^{1}$\quad \textbf{Bertie Vidgen}$^{1}$  \\ \\
{$^1$ Patronus AI $^2$ Contextual AI $^3$ Stanford University}
}

\begin{document}
\maketitle
\begin{abstract}
\textsc{FinanceBench} is a first-of-its-kind test suite for evaluating the performance of LLMs on open book financial question answering (QA). 
It comprises 10,231 questions about publicly traded companies, with corresponding answers and evidence strings. 
The questions in \mbox{\textsc{FinanceBench}} are ecologically valid and cover a diverse set of scenarios. They are intended to be clear-cut and straightforward to answer to serve as a minimum performance standard. 
We test 16 state of the art model configurations (including GPT-4-Turbo, Llama2 and Claude2, with vector stores and long context prompts) on a sample of 150 cases from \mbox{\textsc{FinanceBench}}, and manually review their answers (n=2,400). The cases are available open-source. 
We show that existing LLMs have clear limitations for financial QA. Notably, GPT-4-Turbo used with a retrieval system incorrectly answered or refused to answer 81\% of questions. 
While augmentation techniques such as using longer context window to feed in relevant evidence improve performance, they are unrealistic for enterprise settings due to increased latency and cannot support larger financial documents.
We find that all models examined exhibit weaknesses, such as hallucinations, that limit their suitability for use by enterprises. 
\end{abstract}

\section{Introduction}
Finance specialists routinely need to find information about companies and industries, summarize and analyze that information, and then reason about it. This time-intensive and difficult work is crucial for making investment decisions, developing financial strategies, and conducting due diligence. 
Large Language Models (LLMs) have the potential to augment and automate labor-intensive parts of financial analysis because of their impressive capabilities in natural language understanding, reasoning, and writing \cite{nori2023capabilities, bubeck2023sparks}. %disrupt the financial services industry by augmenting and automating 
However, a key challenge blocking the financial industry's adoption of LLMs is that there are few ways of evaluating models' performance on finance`-specific tasks. 
And, without rigorous, systematic, and measurable evaluation processes, the industry cannot (1) understand the strengths and weaknesses of models; (2) assess whether they perform well enough to use in high-stakes live settings; and (3) track how their capabilities change over time. 

\begin{figure}
    \includegraphics[width=0.5\textwidth]{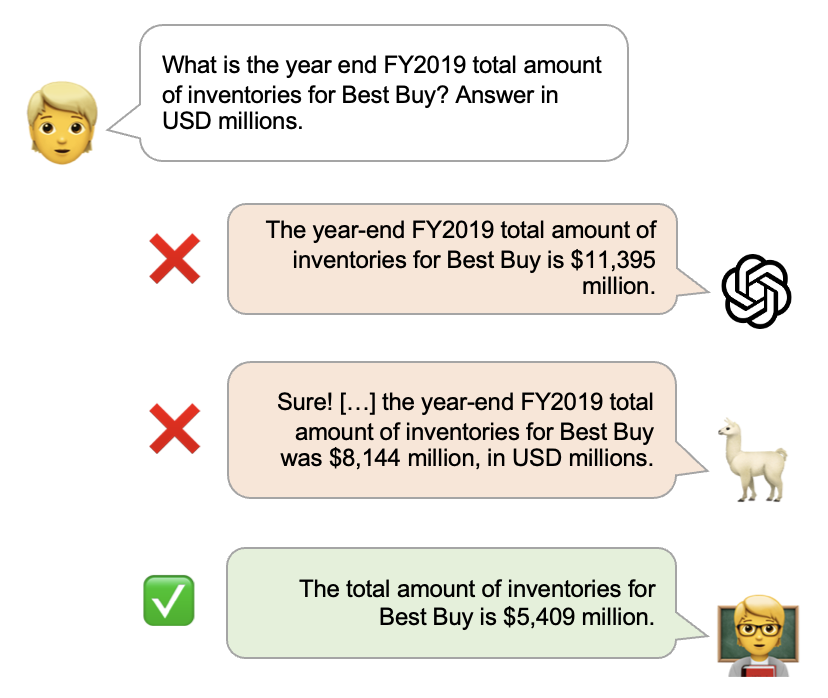}
    \caption{
    Incorrect model responses (using a shared vector store) to a question in \textsc{FinanceBench}. The correct answer is given by the human expert. %The ID is financebench\_id\_01346.
    }
    \label{fig: exaggerated safety example}
\end{figure}

The financial domain presents unique challenges for LLMs. 
First, models need domain-specific knowledge about financial topics and terminology, as well as companies and industries. It is unclear how much financial information and statistics appear in the pre-training data of models. In part to address models' lack of knowledge about finance, BloombergGPT was released in March 2023 as the first LLM specialised for the financial domain \cite{wu2023bloomberggpt}. 
Second, models need up-to-date financial information and to understand relevant financial news. However, many models' data is from several months or years before their release. %As such, most models are unlikely to have extensive and uptodate knowledge of niche financial topics. 
Third, financial questions often involve numerical reasoning. This is a well-established limitation of LLMs, which often make mistakes when asked to make calculations \cite{lu-etal-2023-survey, imani-etal-2023-mathprompter}. 
Fourth, to answer financial questions, models need to handle both unstructured inputs (e.g. qualitative questions in the form of free-text) and structured inputs (such as tabular data) \cite{zhu-etal-2021-tat}. Without additional training, many LLMs are worse at handling tabular inputs than natural language \cite{zha2023tablegpt}.
Fifth, models need to handle multiple bits of information (sometimes from multiple documents) and parse long passages of text. Such content is more difficult for them to reason about than short strings taken from a single source.

To better understand these challenges in using LLMs for Financial QA, we introduce a new benchmark, \textsc{FinanceBench}. %It is a first-of-its-kind benchmark that enables systematic and measurable evaluation of how models perform at answering financial questions. 
It is created by a multidisciplinary team of experts in AI, evaluation, and financial services, and it addresses an important gap in how LLMs are evaluated in finance.
In this paper, we document the construction and composition of \textsc{FinanceBench}, which is intended as an open book test.
We also report the performance of 16 model configurations on \textsc{FinanceBench}, which includes four state of the art models, and a mix of settings (including a closed book, an oracle, two vector store implementations, and a long context window). We provide qualitative insights into their performance. 
From the full \textsc{FinanceBench} dataset, we constructed a diverse sample of 150 cases for evaluation, for which experts manually checked the answers from each of the 16  model configurations models' answers (n=2,400).
The 150 evaluation cases are available open-source.\footnote{\url{https://github.com/patronus-ai/financebench}} 
Data documentation is given in the Appendix, as well as additional information about each of the companies in \textsc{FinanceBench}. 

\section{Prior work}
Several LLMs have been developed for the finance industry, with the release of BloombergGPT attracting considerable attention in early 2023 \cite{wu2023bloomberggpt}. It outperforms other LLMs on generic reasoning benchmarks, financial benchmarks, and proprietary Bloomberg datasets. It is a  50 billion parameter model, trained on 363 million tokens from a proprietary industry-specific dataset, as well as 345 million tokens from generic natural language datasets \cite{wu2023bloomberggpt}. 
Following this work, \citet{yang2023fingpt} introduced FinGPT, an open-source and data-centric model that is trained on a large array of financial data sources. 
\citet{choi2023conversational} created ConFIRM, an LLM-based conversational financial information retrieval model that is designed for financial QA. 
Before the widespread adoption of ChatGPT-style LLMs, \citet{shah-etal-2022-flue} introduced both a new financial language model (``FLANG'') and an evaluation benchmark (``FLUE''), which combines several existing open source financial datasets. FLANG uses preferential financial word and phrase masking during training to improve performance on financial tasks. 
\citet{zhu-etal-2021-tat} created TagOp, BERT- and ELECTRA-based models designed to handle both tabular and textual data. The architecture uses sequence tagging to extract relevant cells from table and text spans, and  symbolic reasoning to derive a final answer. 

Numerous evaluation datasets, benchmarks and test suites have been created that test LLMs' generic capabilities for question-answering, reading comprehension, logical reasoning, and information retrieval \cite{kamalloo-etal-2023-evaluating, qiao-etal-2023-reasoning, huang-chang-2023-towards}. 
They include both ``open book'' tests (where the model has access to external sources of information, such as a document vector store or online sources like Wikipedia) and ``closed book'' tests (where the model has access to no additional information). 
Popular QA benchmarks include SQuAD \cite{rajpurkar-etal-2016-squad} and SQuADRun \cite{rajpurkar-etal-2018-know}, as well as NarrativeQA~\cite{kocisky-etal-2018-narrativeqa} and HellaSwag~\cite{zellers-etal-2019-hellaswag}, which are both part of HELM \cite{liang2023holistic}. 
However, these datasets typically contain no financial questions or only very few (such as TruthfulQA \cite{lin-etal-2022-truthfulqa}). 
And, it cannot be assumed that strong performance on a generic ``open domain'' benchmark generalizes to strong performance in a specific domain, such as financial QA \cite{liu-etal-2022-challenges, niu-etal-2023-learning}. 

FiQA \cite{10.1145/3184558.3192301} was introduced as a shared task to assess how models perform at interpreting financial data, with a focus on aspect-based sentiment analysis and ``opinionated Question Answering''. However, it is limited as sentiment analysis comprises only a small proportion of the questions that financial analysts ask about companies. 
FinQA \cite{chen-etal-2021-finqa} is a high-quality open-source dataset of over 8,000 question and answer pairs, written by financial experts. 
\citet{chen-etal-2022-convfinqa} built on FinQA with ConvFinQA, introduced in 2022. Instead of stand-alone questions, each interaction can involve several questions that may depend on the previous questions/answers. This is a more realistic and more complex testing setup. ConvFinQA comprises 3,892 conversations with 14,115 questions.  
\citet{zhu-etal-2021-tat} introduce TAT-QA, which comprises numerical reasoning questions with tabular and textual data, taken from public financial reports. They used 182 financial reports to construct 16,552 question-answer pairs.
Other benchmarks and tests have been proposed that are finance-specific but are not solely focused on traditional QA. 
\citet{salinas-alvarado-etal-2015-domain} introduced a dataset for named entity recognition of credit risk attributes in financial documents. 
In 2023, \citet{callanan2023gpt} tested whether an LLM could answer mock exam questions for the Chartered Financial Analyst (CFA) Program, levels I and II. 
Although the exact passing criteria for the CFA are not available publicly, the authors estimate that their best-performing implementations would have a ``decent chance of passing''.

\looseness=-1Existing evaluation datasets and tests are not sufficiently grounded in the day-to-day activities of financial analysts. 
They do not address the type of tasks (specifically retrieving information from relevant documents and reasoning about it) that are now being replaced, or substantially augmented, by LLMs. 
This presents a clear risk to ecological validity, and therefore the datasets' usefulness as benchmarks \cite{devries2020ecologically}. 
Therefore, it is critical that LLMs are tested for financial QA with an open-book setup, which involves a clear retrieval component -- rather than just giving them the information that they need to reach the correct answer.

\input{tables/table2} 

\section{\textsc{FinanceBench} Dataset}
\textsc{FinanceBench} is a benchmark dataset that comprises 10,231 questions, answers, and evidence triplets. 
It covers 40 companies that are publicly traded in the USA and 361 public filings, released between 2015 and 2023, including 10Ks, 10Qs, 8Ks, and Earnings Reports. 
Each entry in \textsc{FinanceBench} includes the question itself (e.g. ``What is Boeing's FY2022 cost of goods sold (in USD millions)? ''), the answer (e.g. ``\$63,078 million''), an evidence string (which contains the information needed to verify that the answer to the question is correct) and a page number from the relevant document. 
In some cases, annotators provided a ``Justification'' which explains how they calculated a specific number or reached a conclusion. It was at their discretion to decide whether this field was needed. 
Each entry also has labels for the company name, company's GICS sector, document name, document year, and document type, to enable fine-grained analyses. 
There are three types of questions in \textsc{FinanceBench}.

\looseness=-1First, there are 25 ``\textbf{domain-relevant questions}''. These questions are generically relevant to financial analysis of a publicly-traded company, such as whether it has paid a dividend in the last year, or whether operating margins are consistent throughout multiple financial periods.
The questions were developed with our team of financial analysts and refined by reviewing companies' public findings (e.g. 10Ks) and interviewing financial experts.
In some cases, the questions were not relevant to the company, such as asking about inventory for a technology company or gross margins for a financial services company. In these cases, annotators stated this and gave a brief explanation. 
The domain-relevant questions were posed for 37 of the 40 companies in \textsc{FinanceBench}, contributing 925 entries.

Second, we tasked annotators with creating new questions. They are each specific to the company, the report, and the industry, which we call ``\textbf{novel generated questions}''. 
Annotators were directed to use their knowledge and experience to ask questions that are realistic (in the sense that they relate to important information a financial analyst would want to know); varied (in the sense that they should utilize different parts of the reports, cover different topics, and are phrased differently); and challenging (in the sense that they should not be purely extractive but, instead, involve reasoning). %Annotators were aware of the domain-relevant questions, and directed to ask questions that were different. 
We emphasized ecological validity at all times as we did not want to create a dataset that contains ``challenging'' questions which would not be asked in a real-world setting. 
The novel generated questions were posed to 37 of the 40 companies in \textsc{FinanceBench}. There are between 15 and 80 questions for each company, with an average of 36 questions. 1,323 novel generated questions were created in total. 

Third, we created ``\textbf{metrics-generated questions}''. These are critically important given that a core part of financial analysts' work is to compute metrics and then reason about them. 
Annotators extracted 18 specific metrics ("base metrics") from the three main financial statements in 10Ks (income statement, balance sheet, and cash flow statement), from a period of 8 years (2015-2022). These metrics are mostly standard metrics that many companies report. The base metrics were extracted only if they could be computed using information only within a single financial statement. In other words, if one or more line items within the financial statement clearly represented the metric in question we added the metric into our base metric set. We typically collected 14 metrics per filing as some metrics were either unavailable or ambiguous. 
The base metrics were then used programmatically to construct a series of derivative metrics (metrics whose values are derived from the base metrics). For example, net income margin is derived from the two base metrics: (1) net income and (2) total revenue. We then constructed questions and answers from both the base and the derivative metrics, using templates that were specific to each combination of metric, company, fiscal year, and financial statement(s). %All base and derivative metrics have substantial relevance from the perspective of financially analyzing a company or financially modeling a company.
In some cases, the questions were purely extractive (e.g. ``What is the FY2019 unadjusted operating income (as reported by management) for Amazon?'') and in other cases they involved additional calculations, involving either one or multiple financial statements (e.g. ``what is PepsiCo's FY2021 total D\&A (as shown in cash flow statement) as a percent of total revenue?'').
To ensure that the metrics-generated questions are realistic and varied, the question templates introduced phrasing variations for each of the questions. See details in the Appendix.
The metrics-generated questions were posed to 32 of the 40 companies in \textsc{FinanceBench}. There are between 135 and 348 questions for each company, with an average of 249 questions. 7,983 metrics-generated questions were created in total.

\subsection{Taxonomy of financial questions}
We developed a taxonomy of financial questions, based on taxonomies in prior work \cite{10.1145/3560260} and adapted for the financial services domain. 
We created the taxonomy to better understand the strengths and weaknesses of AI QA tools when addressing different types of questions. 
There are three types of questions in the taxonomy.  
Information extraction refers to extracting specific data or textual content from the filings. Note that the other three types \textit{always} involve some degree of extraction in order to have the information for reasoning.
Numerical reasoning refers to performing mathematical calculations or comparing numerical data.
Logical reasoning refers to using logical deductions to evaluate, contrast, or make judgments regarding the information in the filings. It includes qualitatively assessing information about the company and assessing numerical calculations, such as evaluating computed values.
We applied the questions taxonomy to all of the domain-relevant questions and the metrics-generated questions (total n = 8,908). 
2,493 questions solely involve extracting information (28\%), 5,897 questions involve numerical reasoning (66\%), and 518 (6\%) involve logical reasoning. 
For the metrics-generated questions that involve numerical reasoning (n=5,786), we created a secondary taxonomy label for whether they (1) can be answered with a single financial statement or (2) require multiple financial statements to answer. 
Taxonomy labels are available for the 150 cases in open-source evaluation set.

\subsection{Dataset labelling and quality control}
A team of 20 annotators were recruited for \textsc{FinanceBench}. To join the project, annotators needed to have relevant experience and education in finance, complete a short screening test, and discuss the project with the authors. 
Many were treasury analysts, finance MBAs, and junior analysts. 
Analysts were trained and given access to onboarding and guidance documentation. 
During the early stages of the project, after training had been completed, five annotators left the project due to quality issues and their annotations were discarded.
13 annotators contributed between 19 and 369 of the domain-relevant and novel generated questions. 
2 annotators solely extracted metrics for the metrics-generated questions. They each extracted just over 2,300 metrics, which we used to create 7,983 questions. 
A senior analyst organized, reviewed and gave feedback on the work of the 15 analysts. This analyst has extensive experience in both finance and machine learning, and understood the goals and requirements of the project. The project was run over several weeks, with work issued incrementally as annotators' confidence and experience increased. Each week, approximately 20-25 cases were checked for each annotator (around 10-20\%). Errors were corrected and feedback given to each annotator. In the final stages of the project, we worked with only four annotators who had demonstrated a strong understanding of the task and the quality expectations. 
At the end of the project, approximately 10\% of the domain-relevant and novel generated questions were reviewed and adjustments made to fix quality issues. 
Our analysis of the evaluation samples (see below) indicates that overall the dataset is robust, ecologically valid, and accurate.

\subsection{Dataset for human eval (n=150)}
We created a dataset of 150 cases for human evaluation. It comprises 50 cases from the domain-relevant questions (stratified so there are an equal number of cases from each of the 25 unique questions), 50 randomly sampled novel-generated questions, and 50 randomly sampled metrics-generated questions. 
We sampled evenly from the three types of questions in \textsc{FinanceBench}, despite their different overall volumes, to create a diverse sample that enables fine-grained analysis of model capabilities. This is informed by recent work on the limitations of random sampling for constructing evaluation datasets \cite{vivek2023anchor}. 
We did not stratify the sample by company, year, or industry.

\section{Experimental Setup}
We test four LLMs, from three model providers: OpenAI's GPT-4 \cite{openai2023gpt4} and GPT-4-Turbo with a 128k context window\footnote{\url{https://platform.openai.com/docs/models/gpt-4-and-gpt-4-turbo}}; Anthropic's Claude2 with a 100k context window\cite{bai2022constitutional}; and Meta's Llama2 \cite{touvron2023llama}. We use Llama2 as it is one of the highest-performing open-source models available. 
The LLMs are tested across five setups and two prompt orders (described below) which, because we do not cross every LLM with every setup and prompt format, results in 16 distinct configurations. 
The configurations test either an important (albeit artificial) LLM implementation (i.e. the Closed book and the Oracle settings) or an implementation that reflects how LLMs are being adopted in industry for Financial QA. %The implementation details are described in the Appendix. 
The prompt templates are described in the Appendix. 

\paragraph{Closed book}
For GPT-4 and GPT-4-Turbo, we test a closed book setting.
Each prompt is fed to the model without any additional information or context.\footnote{During our early testing, we compared GPT-4 against GPT-3.5. GPT3.5's performance was similar but slightly worse. Due to this, we decided not to continue testing GPT-3.5 further.} This is the most naive implementation of an LLM for financial QA. 

\paragraph{Oracle}
We also test an unrealistic Oracle setting for both GPT-4 and GPT-4-Turbo. In this setting, the model is given the prompt as well as the text from the page used to evidence the answer (as recorded by the annotators during  dataset creation). In principle, all of the information that it needs to answer the question. This turns the task into ``open book'' question answering by removing the retrieval challenge, which makes it both unrealistic and substantially easier.
For all non metrics-generated questions, we used the entire page text from the same page(s) as the evidence texts that annotators labelled (therefore the model had full page context around the specific evidence text that annotators chose to answer the question at hand). We added the relevant page(s) to each prompt before feeding it to the model. 
For the metrics-generated questions, we provided the relevant financial statement(s) from the document needed to calculate the metric, such as the cash flow statement and / or income statement.
We present these results solely as a reference study.

\paragraph{Single vector store}
We create a simple retrieval baseline by initializing a single vector store per document. While this is unrealistic in production settings, we construct this as a naive baseline. Vector stores enable models to quickly access, and use, relevant information for a given task, and have been proposed as a way of making AI models more factually- and contextually- grounded~\cite{10.5555/3495724.3496517}. A single vector store setup is slower to run as we have to construct a new vector store each time, and is arguably unrealistic in a live industry setting where thousands of documents are available for use. However, it means the vector store has a much smaller range of documents to search over and so should perform better. We test GPT-4, GPT-4-Turbo, and Llama2 with the single vector store. 

\paragraph{Shared vector store}
We construct a more realistic setting by creating a shared vector store for all documents. We use the same Chroma database, Langchain implementation and OpenAI embeddings as for the single vector store. Our vector store is implemented in Langchain\footnote{\url{https://www.langchain.com/}}, using a Chroma database\footnote{\url{https://www.trychroma.com/}} and OpenAI embeddings (text-embedding-ada-002). The vector store indexes over all of the 360 documents that appear in FinanceBench. 
We test GPT-4, GPT-4-Turbo, and Llama2 with the shared vector store. 

\paragraph{Long context}
GPT-4-Turbo and Claude2 are capable of handling long context windows (128k and 100k tokens, respectively). The public filing that the question relates to is added to the prompt and then fed in to the model. 
This removes the need for a vector store, and offers a flexible way of handling documents. 
In some cases, the long context window is still not sufficient to handle the public filing (which can run to 250 pages). For these cases, we truncated the document to the first 95,000 to 100,000 tokens. This choice is partly justified by the fact that nearly all questions relate to the earlier parts of the documents. 
Also note that long context windows today are not only too small to support documents typically used by financial analysts, such LLMs are also much slower and more expensive to use. Therefore, they are not typically used in a production setting today.

\paragraph{Prompt order: Relative position of question and context}
For all setups that involve passing the model relevant information (i.e. every setting apart from the closed book), the order of the prompt and the evidence string can be swapped round; the prompt can go before or after the evidence. This can make a substantial difference to how models perform, especially with longer evidence strings. We refer to these two prompt schemes as Context-First or Context-Last. 
We test both the GPT-4 and GPT-4-Turbo on oracle settings, and the GPT-4-Turbo and Claude2 on long context settings with both prompt schemes.

\subsection{Labelling LLM Responses}
\label{sec:dataset_human}
Each of the models' responses to the 150 questions have been labelled by one of the research team. Complex cases were raised for discussion, and samples from every models' responses were spot-checked. %Every label was checked by at least one other person, and complex cases were raised for discussion.
Models' responses were each assigned to one of three categories. 
First, \textbf{correct answer}. This is the `desired' behavior of models. To ensure a good-faith understanding of models' capabilities we allow minor deviations, such as giving the answer in billions when the unit was given in the question as millions. We also allow very small rounding errors. 
Second, \textbf{incorrect answer}. Incorrect answers vary, from calculations that are off by small margins to several orders of magnitude, and from making up legal information to giving the wrong direction for an effect (e.g. reporting negative growth when it is actually positive). 

If a model gives the right answer but with logic or calculations that explicitly contradict the evidence in the gold standard answer, we label it Incorrect. 
Third, \textbf{failure to answer}. If the model explicitly states that it cannot answer because it does not have access to the right information then it is a failure to answer (e.g. ``As an AI, I don't have real-time data access capabilities to provide information on Boeing's production rate forecast for FY2023.'').
\section{Results on \textsc{FinanceBench}}

\input{tables/table4}

\begin{figure*}
    \vspace{-5mm}
    \centering
    \includegraphics[width=1\textwidth]{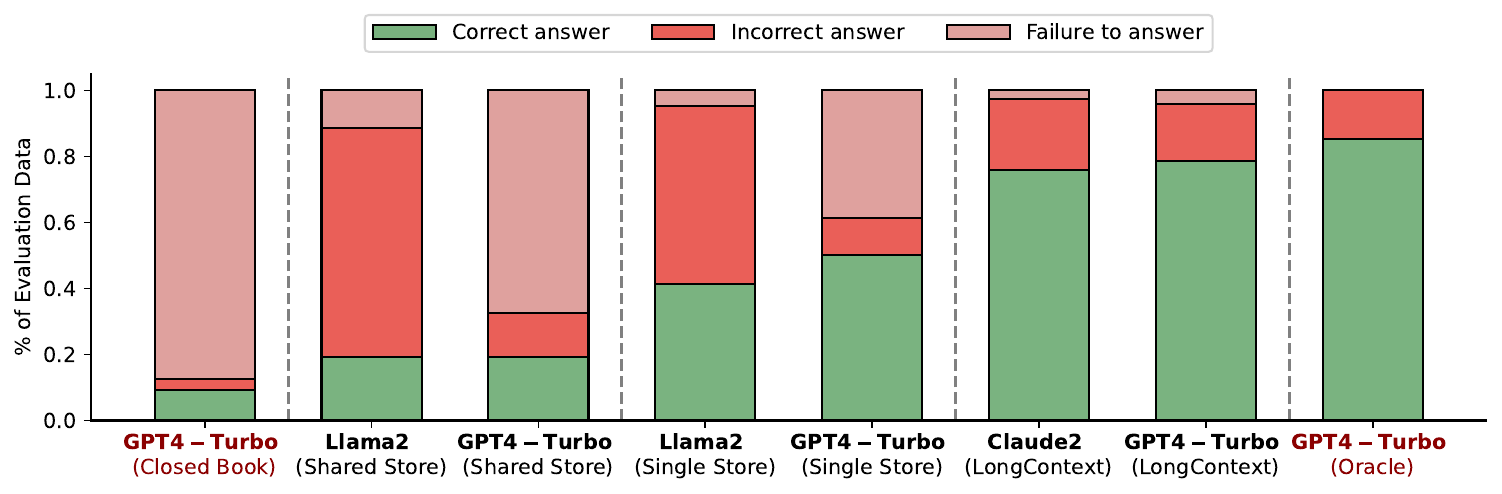}
    \caption{
    Performance of 8 model configurations on \textsc{FinanceBench} human eval sample (n=150). The Oracle setting and the Closed Book setting are highlighted in red as these represent unrealistic evaluation scenarios that only serve as reference points. 
    }
    \label{fig: financebench-overall}
\end{figure*}

\paragraph{Overall performance}
Without access to additional information (i.e. in a closed book configuration), models perform poorly on \textsc{FinanceBench}. GPT-4-Turbo (Closed Book) only gives correct answers to 9\% of prompts \footnote{\;While we have evaluated GPT-4 and GPT-4-Turbo, we only show the better performing model GPT-4-Turbo in the main text. A detailed comparison is provided in the Appendix.}. 
Augmentation techniques, such as incorporating public filings in a long-context window and using a vector store, vary in how effective they are, depending partly on how they are implemented (see Figure~\ref{fig: financebench-overall}), with success rates from 20\% to 78\%. A correct answer is considered a success. 
The Oracle (GPT-4-Turbo with evidence pages) is 85\% successful.
As anticipated, the configuration of GPT-4-Turbo with one vector store for \textbf{each document} had a higher success rate than the configuration with a single vector store for \textbf{all documents} (50\% vs 19\%). We observe the same trend for Llama2 (41\% vs 19\%). However, the models exhibited different weaknesses. Llama2 had a much higher percentage of incorrect answers (70\% and 54\%) rather than failing to answer (11\% and 5\%) whereas the equivalent GPT-4-Turbo had far more failure to answers (67\% and 38\%) than incorrect answers (13\% and 11\%).

Anthropic's Claude-2 with long-context success rate is 76\% and OpenAI's GPT-4-Turbo with long context is 79\%. Like the Llama2 vector-store configuration, these two models had far more incorrect answers (21\% and 17\%) than refusals (3\% and 4\%). 
In an industry setting, the high proportion of failures which are incorrect answers rather than refusals could be still concerning as it indicates a greater risk of hallucinations. Models refusing to answer is arguably preferable to giving an incorrect answer as it creates less risk of error, and misplaced trust, by users. 
Overall, these findings indicate that (1) access to the right information (i.e. a vector store or similar) and (2) correct information retrieval is critical for models to perform well at financial QA. However, once the right information has been extracted, they still need to reason correctly -- and models still demonstrate weaknesses in this regard. 

\begin{figure}[htb!]
    \small
    \centering
    \includegraphics[width=0.47\textwidth]{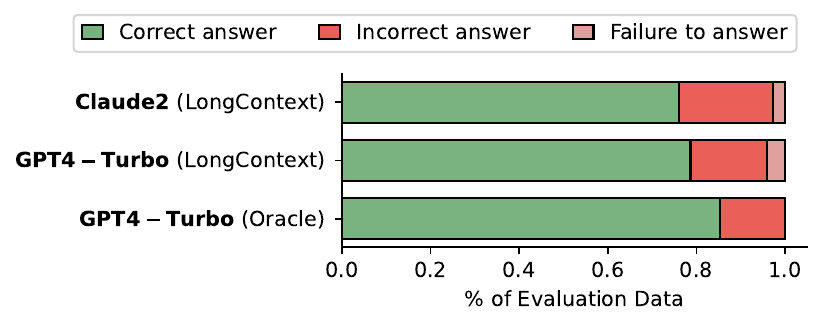}\\
    (i) Context-First Prompting Scheme (Default)
    \includegraphics[width=0.47\textwidth]{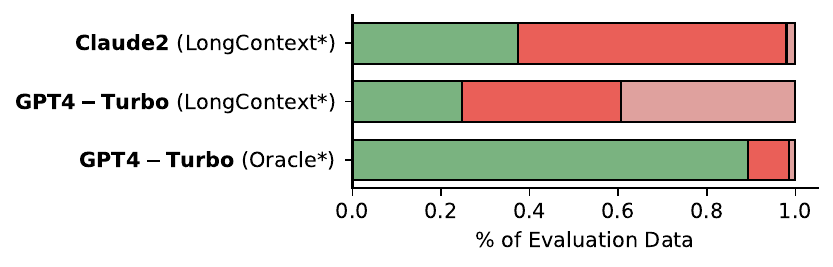}\\
    (ii) Context-Last Prompting Scheme
    \caption{Ablation study of different prompting schemes on \textsc{FinanceBench} human eval sample (n=150). Showing the relevant context (i.e., filing or evidence extract) before the question leads to significant performance improvements over showing the context after the question.}
    \label{fig:prompt_order_ablation}
\end{figure}

\begin{figure*}[!htb]
    \vspace{-5mm}
    \centering
    \includegraphics[width=1\textwidth]{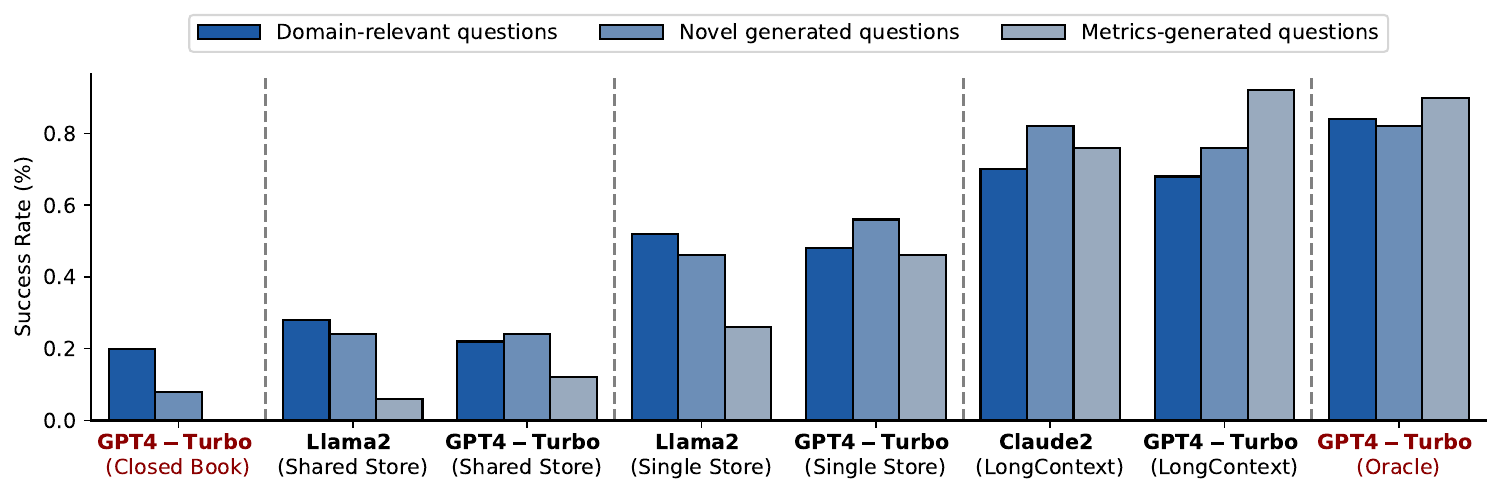}
    \caption{Performance of eight model configurations on \textsc{FinanceBench} human eval sample (n=150) by type of question. The Oracle setting and the Closed Book setting are highlighted in red as these represent unrealistic evaluation scenarios that only serve as reference points. }
    \label{fig: Model performance on FinanceBench by type of question}
\end{figure*}

\paragraph{Performance by question type} Models' performance varies across the three types of questions in \textsc{FinanceBench} (see Figure ~\ref{fig: Model performance on FinanceBench by type of question}. 
Models typically perform worst on the metrics-generated questions, apart from the long context setting where Claude2 performs second best and GPT-4-Turbo performs the best. This suggests that part of the challenge with metrics-generated questions is retrieving the correct information. 
We reviewed the evaluation dataset in-depth, and many of the freely generated questions only involve extraction. It is therefore unsurprising that models perform better on these questions. 
Equally, many of the domain-relevant could be answered by using general world knowledge. For instance, they cover what industry a company operates in and what their main services and products are. 
In contrast, many of the metrics-generated questions involve more complex numeric reasoning and require using multiple passages from the documents. 

\paragraph{Performance by prompt order} The relative order of the relevant context and the question of interest has a clear impact on the models performances in the LongContext setting (see Figure~\ref{fig:prompt_order_ablation}). Presenting the relevant filing first and then appending the question of interest (Context-First scheme) leads to significantly improved sucess rates for both GPT-4-Turbo (78\% vs. 25\%) and Claude2 (76\% vs. 37\%) in the LongContext setting. Surprisingly, we cannot observe the same trend in the Oracle setting where the provided context is of significantly shorter length as it is only of the form of an evidence extract (e.g.,\ one to few pages). The reversed prompt order (i.e.,\ Context-Last) leads to slightly better performance (89\% vs. 85\%) in this setting. We hypothesize that the strong performance difference in the LongContext setting stems from models loosing track of the question of interest after seeing thousands of evidence tokens in the in the Context-Last prompting scheme.

\subsection{Qualitative analysis of responses}
As well as labelling the models' responses to the 150 evaluation cases with the three labels (Correct answer, Incorrect answer, and Refusal to Answer), we also qualitatively analyze models' responses to identify patterns and themes. We grouped them together into the following five themes. 

\paragraph{High-quality correct answers} In some cases, models gave high-quality, fully-evidenced, and fully comprehensible correct answers, which are more useful and informative than the gold standard answers. 
This includes providing multiple bits of evidence, calculating both absolute and percentage differences, and clearly explaining each step of a calculation. 
When correct, the responses from Anthropic's Claude-2 and OpenAI's GPT-4-Turbo long-context window setup were often particularly high-quality.

\paragraph{Different but valid correct answers} Some answers are substantively different to the gold standard answers, but still valid.
This is often the case with more qualitative assessments, where models provide a reasonable and informative explanation which differs from the gold standard. 
In some cases, the test cases do not specify units or the type of evidence that is needed (e.g. for cases that involve assessing whether a company is ``capital-intensive''). This leads to ambiguity where several different bits of evidence could be given to substantiate a position. 
To ensure a good-faith assessment of models' answers, we consider different but valid answers to be correct. 

\paragraph{Hallucinations} 
In many cases, models gave superficially coherent and seemingly well-justified answers, sometimes with extensive calculations and reasoning steps, which were still wrong. We consider these as ``hallucinations'' because they are a response where the generated output is unfaithful to the given source \cite{@10.1145/3583780.3614905}. These are particularly concerning as they are harder to catch.%  because the model is so confident. 
The Llama2 configurations are more likely to give plausible but incorrect answers than the GPT-4-Turbo configurations, as evidenced by the higher percentage of responses that are an Incorrect answer than a Failure to answer. 

\paragraph{Helpful refusals} In some cases, particularly for the closed book setting and the vector-store setups, models refuse to answer the test case but still explain \textit{how} it could be answered, such as by giving advice on where to find relevant information. Models would also provide general information about the company, the metric, or financial analysis in general. This is a useful response, but technically still a failure.

\paragraph{Irrelevant comments} In some cases, models' responses did not address the question. This indicates that they do not properly understand the task, and there is a large element of ``guessing''.

\section{Limitations of \textsc{FinanceBench}}

\paragraph{Single-turn conversations} 
\textsc{FinanceBench} contains only single questions and answers. 
However, financial analysts often ask a stream of questions within a single conversation so they can dig deeper into a single industry, company, or topic. They also ask questions dynamically; adjusting questions if the model gives an inadequate response or, alternatively, asking additional questions if the response is high-quality and it spurs followups, such as explaining a metric or providing more information. Nonetheless, the primary usecase, and biggest priority, is for LLMs to provide high-quality responses to single questions. %Single-turn conversations are also the ``starting point'' before multi-turn conversations can be addressed.

\paragraph{Public filings} 10Ks, 10Qs, 8Ks, and Public earnings reports are key documents used by financial analysts to assess companies and industries, and to make decisions. 
However, some analysts also use documents that are not in the public domain. This is particularly common with venture capital, where companies are typically not publicly listed. We focused on only public filings because private documents create issues around commercial sensitivity and privacy. %; and any documents that we had access to could not be shared with others, which would limit the accessibility and useability of \textsc{FinanceBench}. 
Equally, another limitation of \textsc{FinanceBench} is that it only contains publically listed companies, which necessarily biases the dataset towards larger companies and properly-audited and well-written documents.

\paragraph{Lack of cross-company comparisons} \textsc{FinanceBench} was designed to answer questions about a single company, rather than to compare figures between two companies. This is partly because our interviews showed that analysts primarily ask companies about single companies; and partly because comparing two companies means that models have to handle two separate documents, which is much harder than handling even two strings from a single document \cite{yang-etal-2018-hotpotqa}. %Given the performance of  QA models, we would anticipate a very low success rate. %In a sense, this is not a limitation of the dataset as a limitation of the tools -- and we did not want to test them against a task that they would be unable to complete.

\paragraph{Dataset integrity} There are two main limitations to the quality of the dataset. 
First, some of the questions are ecologically valid but simplistic. This makes them suitable for a first line of evaluation, but it means that most models can answer them correctly (often without using any additional information source), which leads to higher performance on the benchmark. % and therefore many models can answer them correctly. 
Second, sometimes the correct answer is ambiguous. It can depend on the context and assumptions/priorities of the analyst. This means that some gold standard labels are valid but still contestable. 
Overall, from our extensive reviews and analysis of the dataset, we believe that the gold labels are high quality.

\section{Conclusion}
\textsc{FinanceBench} reveals critical weaknesses in the performance of state of the art models at financial QA. 
Several of the models we tested had critical weaknesses. 
Outside of the unrealistic Oracle setting, even the very best performing model that we test (GPT-4-Turbo with the long-context window) is still only correct in 79\% of cases. Such a model could not be used with confidence in a live industry setting. 
And, concerningly, even if an LLM appears to be giving reasonable responses, there remains a risk that its answers are hallucinations, out-of-date, logically incorrect, or given with the wrong units. All are serious risks to effective financial analysis, and may not be apparent without detailed inspection of the results. 

Given the limitations identified by \textsc{FinanceBench} we encourage all analysts using AI for financial QA to (1) robustly evaluate their models before using them in high-stakes live settings; (2) use additional sources of information (such as vector stores and long-context content) to improve performance; and (3) double check results and triangulate findings by using multiple sources of evidence. 
We also encourage other researchers to build on our findings and test other AI models and retrieval systems, as well as approaches such as fine-tuning, few-shot learning, chain of thought \cite{wang-etal-2023-towards}, and adding additional ``tools'' such as calculators and APIs, could drive better performance. 
Future work will expand the scope and coverage of \textsc{FinanceBench} and address the limitations identified in this paper.

\bibliography{custom}
\bibliographystyle{acl_natbib}

\clearpage
\appendix

%%%%%%%%%%%%%%%%%%%%%%%%%%%%%%%%%%%%%%%%%%%%%%%%%%%%%%%%%%%%%%%%%%%%%%%%%%%%%%%%%%%%%%%%%
%%%%%%%%%%%%%%%%%%%%%%%%%%%%%%%%%%%%%%%%%%%%%%%%%%%%%%%%%%%%%%%%%%%%%%%%%%%%%%%%%%%%%%%%%%
\section{Phrasing variations for the metrics-based questions}
To ensure that the metrics-generated questions are realistic and diverse, we used templates to create phrasing variations for each of the base questions. 
This includes 11 ``vanilla'' introductory clauses and 11 more creative introductory clauses; and 7 vanilla and 7 more creative ending clauses; as well as 2-3 unique ways of referring to each financial statement. We also introduced randomness in the ordering that statements are referred to, when multiple are referenced, and randomness in the units. For instance, the examples just given are in the benchmark as ``What is the FY2019 unadjusted operating income (as reported by management) for Amazon? Answer in USD millions. Please utilize information provided primarily within the income statement.'' and ``We want to calculate a financial metric. Please help us compute it by basing your answers off of the income statement and the cash flow statement. Here's the question: what is PepsiCo's FY2021 total D\&A (as shown in cash flow statement) as a percent of total revenue?''.

\begin{table*}[]
    \centering
    \resizebox{0.9\textwidth}{!}{%
    \begin{tabular}{lc|l}
        \toprule
         \textbf{Configuration} & \textbf{Prompt Order} &  \textbf{Prompt}  \\
         \toprule
          Closed Book & - & Answer this question: {\color{blue}\texttt{[QUESTION]}}\\
          \midrule
          Shared Vector Store & - & Answer this question: {\color{blue}\texttt{[QUESTION]}}\\
          \midrule
          Single Vector Store & - & Answer this question: {\color{blue}\texttt{[QUESTION]}}\\
          \midrule
          \multirow{6}{*}{LongContext} & \multirow{3}{*}{Context-First} & Answer this question: {\color{blue}\texttt{[QUESTION]}} \\
          & & Context: \\
          & & [START OF FILING] {\color{blue}\texttt{[FILING]}} [END OF FILING]\\
          \cmidrule{2-3}
          & \multirow{3}{*}{Context-Last}  & Context: \\ 
          & &[START OF FILING]{\color{blue}\texttt{[FILING]}} [END OF FILING] \\
          & & Answer the following question: {\color{blue}\texttt{[QUESTION]}} \\
          \midrule
          \multirow{6}{*}{Oracle} & \multirow{3}{*}{Context-First} & Answer this question: {\color{blue}\texttt{[QUESTION]}} \\
          & & Context: \\
          & & [START OF FILING] {\color{blue}\texttt{[EVIDENCE EXTRACT]}} [END OF FILING]\\
          \cmidrule{2-3}
          & \multirow{3}{*}{Context-Last}  & Context: \\ 
          & &[START OF FILING]{\color{blue}\texttt{[EVIDENCE EXTRACT]}} [END OF FILING] \\
          & & Answer the following question: {\color{blue}\texttt{[QUESTION]}} \\
         \bottomrule
    \end{tabular}
    }
    \caption{Prompt setups for the different evaluation configurations. Blue colored text denotes placeholder for the question of interest and the relevant context. In addition, we added start- and end-delimiters ([START OF FILING] and [END OF FILING]) to the prompts in the LongContext and Oracle configurations.}
    \label{tab:prompt_setups}
\end{table*}

\section{LLM implementation}
\label{app: llm-implementation}
We test 16 model configurations, including the Oracle and Closed book settings. 
All were tested in November 2023. 
Llama2 was accessed through Replicate\footnote{\url{https://replicate.com/meta/llama-2-70b-chat}} and the OpenAI and Anthropic models were accessed through their respective APIs. We used the default system prompts for all calls. Temperature was set to 0.01 and max token length to 2,048. We show the employed prompts in Table~\ref{tab:prompt_setups}. For the long context configurations, we truncated the filing input to 95,000 tokens if the relevant filing was not fitting into the possible context.

\section{Data Documentation}
\textsc{FinanceBench} comprises 10,231 cases, of which 150 are used for expert evaluation and are available open-source. 

\paragraph{Summary of columns}
There are 16 columns in the dataset, which are associated with every entry. 
\begin{enumerate}
    \item A unique ID (of the form, ``finacebench\_id\_0000'').
    \item A value for whether it is in the eval sample of 298 cases (`1'), in the open source sample (`2') or in neither (`0'). 
    \item The company's name.
    \item The company's sector following GICS sector definitions.
    \item The name of the public filing used to pose and answer the question. 
    \item A link to the relevant public filing. Where possible, we used static PDFs from the company's investor relations page or other reputable sources like EDGAR.
    \item A label for the document type (e.g. 10K, 10Q).
    \item The fiscal year that the document is referencing. If the document is an 8K, then it refers to calendar year since these documents generally are not released following fiscal year calendars. The fiscal years were labelled using the following convention: use the calendar year of the fiscal year end as the fiscal year. This means if a fiscal year ends in January 2023, we label that fiscal year as FY2023. The one exception to this rule is Johnson \& Johnson whose fiscal year ends in the first few days of January. 
    \item The question type (reflecting the three types in \textsc{FinanceBench}: domain-relevant, novel-generated, and metrics-generated).
    \item The type of reasoning (e.g. numerical reasoning).
    \item The domain-relevant question number (if relevant), which runs from dg01 to dg25. 
    \item The actual question.
    \item The gold standard answer. 
    \item The evidence text. In the cases of domain-relevant questions and novel-generated questions, these are the evidence texts that annotators directly extracted themselves. In the case of metrics-generated questions, we constructed the evidence text as follows: (i) for each base metric that is a building block of the main metric in question, extract the page number from the PDF where that base metric was calculated or extracted; (ii) using the PDF page number, extract the entire PDF page text so as to ensure much (if not all) of the financial statement, where the base metric was found in, is extracted as well; (iii) combine the different full page texts and remove duplicates
    \item The evidence text page number. Note that all page numbers are 1-indexed.
    \item The full page text found in the financial document for each evidence text page number. This is to provide a larger relevant context around each evidence text chosen by annotators.
    \item Where relevant, the justification for each answer.
\end{enumerate}

\paragraph{Dataset description}
There are 40 companies in \textsc{FinanceBench}, of which 32 companies are in the metrics-generated questions and 37 are in the domain-generic and novel-generated questions. 29 companies appear in all three types of questions. 
There are 360 documents in total, of which 270 are 10Ks (the vast majority, accounting for 75\% of all documents), 5 are annual reports (which largely cover the exact same content as the 10k but with additional prose at the start), 29 are 8Ks, 29 are Earnings reports, and 27 are 10Qs. 
The distribution of questions is more unequal, with 10Ks accounting for 9,530 questions (or 93\% of the total). This is due to the technical detail contained within 10Ks, as well as their importance within the finance industry. 
Nine of the 11 GICS sectors are represented, ranging from Information Technology (25\% of questions) to Materials (1.8\%). 
Every entry has an evidence text and evidence page number. 
749 of the domain-relevant and novel-generated questions have justifications. 

\section{Comparison of GPT-4 and GPT-4-Turbo}
\begin{figure*}
    \centering
    \includegraphics[width=1.0\linewidth]{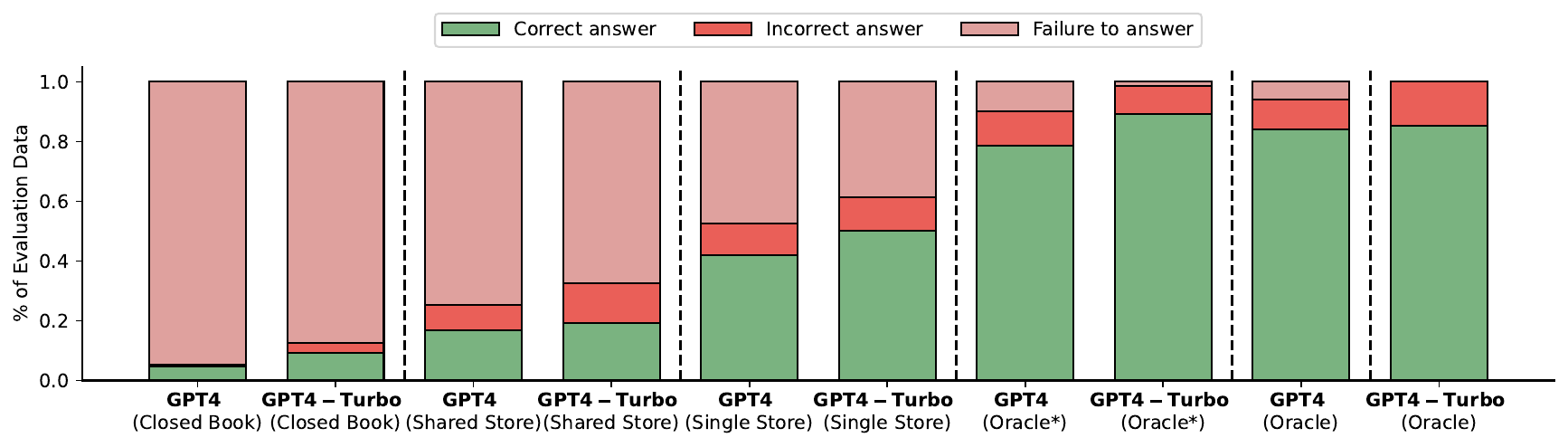}
    \caption{Performance Comparison of OpenAI's GPT-4 and GPT-4-Turbo across the different evaluation configurations. Note that we cannot compare performance on the \texttt{LongContext} setting as GPT4 doesn't support a long context setting.}
    \label{fig:gpt_comparison}
\end{figure*}

While we have evaluated GPT-4 and GPT-4-Turbo, we only show the better performing model GPT-4-Turbo in the main text. We provide a comparison on the model performances in each evaluated setting in Figure \ref{fig:gpt_comparison}. Note that we cannot compare performance on the \texttt{LongContext} setting as GPT-4 doesn't support a long context setting.

\section{Information about each company}
See the information in Table~\ref{tab:company_summary}.
\input{tables/table5}

\end{document}

%% file: tables/table2.tex
\begin{table*}[t]
\vspace{-5mm}
\small
\centering
\begin{tabular}{l|l|ccc|c}
\toprule
\multirow{2}{*}{\parbox{2cm}{\centering\textbf{Company}}} & \multirow{2}{*}{\parbox{4cm}{\centering\textbf{GICS Sector}}} & \multirow{2}{*}{\parbox{1.6cm}{\centering\textbf{Metrics-generated}}} & \multirow{2}{*}{\parbox{1.6cm}{\centering\textbf{Domain-relevant}}}  & \multirow{2}{*}{\parbox{1.6cm}{\centering\textbf{Novel generated}}} & \multirow{2}{*}{\parbox{1.6cm}{\centering\textbf{Total}}} \\
 & & & & \\
\midrule
\rule{0pt}{4pt}% Adjust the height of the space here
3M & Industrials & 269 & 25 & 36 & 330 \\
Boeing & Industrials & 248 & 25 & 47 & 320 \\
CVS Health & Health Care & 253 & 25 & 23 & 301 \\
Coca-Cola & Consumer Staples  & 239 & 25 & 25 & 289 \\
MGM Resorts & Consumer Discretionary & 196 & 25 & 26 & 247 \\
Netflix & Communication Services & 242 & 25 & 25 & 292 \\
Pfizer & Health Care & 156 & 25 & 58 & 239 \\
Salesforce & Information Technology & 0 & 25 & 25 & 50 \\
Ulta Beauty & Consumer Discretionary & 0 & 25 & 25 & 50 \\
Verizon & Communication Services & 272 & 25 & 50 & 347 \\
\bottomrule
%\textbf{Total} & & \textbf{7,983} & \textbf{925} & \textbf{1,323} & \textbf{10,231} \\
\end{tabular}
\caption{Selection of 10 companies in \textsc{FinanceBench}}
\label{tab:dataset_summary}
\end{table*}

%% file: tables/table4.tex
\begin{table*}[ht]
\small
\centering
\begin{tabular}{ll|ccc|c}
\toprule
\multirow{2}{*}{\textbf{Model}} & \multirow{2}{*}{\textbf{Configuration}} & \multirow{2}{*}{\textbf{Correct answer}} & \multirow{2}{*}{\textbf{Incorrect answer}} & \multirow{2}{*}{\textbf{Failed to answer}} & \multirow{2}{*}{\textbf{Total}} \\
&&&& \\ % This empty row is necessary for the multirow cells to align properly
\midrule
\multirow{2}{*}{\texttt{GPT-4-Turbo}} & Closed Book    & \multirow{2}{*}{14 (9\%)}      & \multirow{2}{*}{5 (3\%)}  & \multirow{2}{*}{126 (88\%)}  & \multirow{2}{*}{150}            \\
& (on its own) &  &   & &             \\
\midrule
 % \multirow{2}{*}{Llama2}&Shared Vector Store & 29 (19\%)   & 104 (70\%)   & 17 (11\%) & 150  \\
 %    &(one store for all filings) \\
\multirow{2}{*}{\texttt{Llama2}} & Shared Vector Store & \multirow{2}{*}{29 (19\%)}   & \multirow{2}{*}{104 (70\%)}   & \multirow{2}{*}{17 (11\%)} & \multirow{2}{*}{150  }          \\
 & (one store for all filings) &  &   & &             \\
%\midrule
\multirow{2}{*}{\texttt{GPT-4-Turbo}} & Shared Vector Store & \multirow{2}{*}{29 (19\%)}  & \multirow{2}{*}{20 (13\%)} & \multirow{2}{*}{101 (68\%)} & \multirow{2}{*}{150}            \\
& (one store for all filings) &  &   & &             \\
\midrule
\multirow{2}{*}{\texttt{Llama2}} & Single Vector Store & \multirow{2}{*}{62 (41\%) }  & \multirow{2}{*}{81 (54\%) }   & \multirow{2}{*}{7 (5\%)} & \multirow{2}{*}{150 }           \\
& (one store for each filing) &  &   & &             \\
%\midrule
%GPT-4 with evidence pages
\multirow{2}{*}{\texttt{GPT-4-Turbo}} & Single Vector Store & \multirow{2}{*}{75 (50\%)} & \multirow{2}{*}{17 (11\%)} & \multirow{2}{*}{58 (39\%)}  & \multirow{2}{*}{150  }         \\
& (one store for each filing) &  &   & &             \\
\midrule
\multirow{2}{*}{\texttt{Claude2}} & Long Context & \multirow{2}{*}{114 (76\%)}   & \multirow{2}{*}{32 (21\%)} & \multirow{2}{*}{4 (3\%)} & \multirow{2}{*}{150 }           \\
& (filing in context) &  &   & &             \\
%\midrule
\multirow{2}{*}{\texttt{GPT-4-Turbo}} & Long Context & \multirow{2}{*}{118 (79\%)}   & \multirow{2}{*}{26 (17\%)} & \multirow{2}{*}{6 (4\%)}  & \multirow{2}{*}{150 }           \\
& (filing in context) &  &   & &             \\
\midrule
\multirow{2}{*}{\texttt{GPT-4-Turbo}} & Oracle& \multirow{2}{*}{128 (85\%)}  &\multirow{2}{*}{ 22 (15\%)} & \multirow{2}{*}{0 (0\%)}  &\multirow{2}{*}{ 150 }           \\
& (access to evidence pages) &  &   & &             \\
\midrule
% \textbf{Total} & 733 & 407 & 648 & 1,788 \
\multirow{2}{*}{\textbf{Total}} & & \multirow{2}{*}{569 (47\%)} & \multirow{2}{*}{307 (26\%)} & \multirow{2}{*}{324 (27\%)} & \multirow{2}{*}{1200} \\ % Empty cell for the second line of the total row
&&&& \\ % Empty row for the second line of the total row\
\bottomrule
\end{tabular}
\caption{Model performance of 8 model configurations on \textsc{FinanceBench} human eval sample (n=150).}
\label{tab:model_performance_table}
\end{table*}

%% file: tables/table5.tex
\begin{table*}[ht]
\centering
\begin{tabular}{l|c|c|l|c}
\toprule
\textbf{Company}     & \textbf{Symbol} & \textbf{Market cap} & \textbf{GICS Sector}   & \textbf{S\&P 500} \\
\midrule
3M                   & MMM             & \$48.5 billion      & Industrials            & Yes               \\
AES Corporation      & AES             & \$8.44 billion      & Utilities              & Yes               \\
Amcor                & AMCR            & \$12.92 billion     & Materials              & Yes               \\
AMD                  & AMD             & \$166.27 billion    & Information Technology & Yes               \\
Activision Blizzard  & ATVI            & \$73.7 billion      & Communication Services & Yes               \\
Adobe                & ADBE            & \$235.14 billion    & Information Technology & Yes               \\
American Express     & AXP             & \$108.33 billion    & Financials             & Yes               \\
American Water Works & AWK             & \$23.11 billion     & Utilities              & Yes               \\
Apple                & AAPL            & \$2730 billion      & Information Technology & Yes               \\
Best Buy             & BBY             & \$14.72 billion     & Consumer Discretionary & Yes               \\
Boeing               & BA              & \$112.37 billion    & Industrials            & Yes               \\
CVS Health           & CVS             & \$89.59 billion     & Health Care            & Yes               \\
Coca-Cola            & KO              & \$226.51 billion    & Consumer Staples       & Yes               \\
Corning              & GLW             & \$25.32 billion     & Information Technology & Yes               \\
Costco               & COST            & \$252.18 billion    & Consumer Staples       & Yes               \\
eBay                 & EBAY            & \$22.68 billion     & Consumer Discretionary & Yes               \\
FedEx                & FDX             & \$65.16 billion     & Industrials            & Yes               \\
Foot Locker          & FL              & \$1.79 billion      & Consumer Discretionary & No                \\
General Mills        & GIS             & \$36.25 billion     & Consumer Staples       & Yes               \\
Intel                & INTC            & \$150.31 billion    & Information Technology & Yes               \\
JPMorgan             & JPM             & \$415.28 billion    & Financials             & Yes               \\
Johnson  Johnson     & JNJ             & \$378.4 billion     & Health Care            & Yes               \\
Lockheed Martin      & LMT             & \$100.07 billion    & Industrials            & Yes               \\
MGM Resorts          & MGM             & \$12.21 billion     & Consumer Discretionary & Yes               \\
McDonalds            & MCD             & \$183.82 billion    & Consumer Discretionary & Yes               \\
Microsoft            & MSFT            & \$2370 billion      & Information Technology & Yes               \\
Nike                 & NKE             & \$146.56 billion    & Consumer Discretionary & Yes               \\
Netflix              & NFLX            & \$165.11 billion    & Communication Services & Yes               \\
Oracle               & ORCL            & \$296.81 billion    & Information Technology & Yes               \\
PGE Corporation      & PCG             & \$39.41 billion     & Utilities              & Yes               \\
Paypal               & PYPL            & \$63.12 billion     & Financials             & Yes               \\
PepsiCo              & PEP             & \$220.39 billion    & Consumer Staples       & Yes               \\
Pfizer               & PFE             & \$188.97 billion    & Health Care            & Yes               \\
Salesforce           & CRM             & \$196.56 billion    & Information Technology & Yes               \\
Ulta Beauty          & ULTA            & \$19.14 billion     & Consumer Discretionary & Yes               \\
Verizon              & VZ              & \$133.77 billion    & Communication Services & Yes               \\
Walmart              & WMT             & \$428.17 billion    & Consumer Staples       & Yes               \\
Block                & SQ              & \$26.8 billion      & Information Technology & No                \\
Amazon               & AMZN            & \$1325 billion      & Consumer Discretionary & Yes               \\
Kraft Heinz          & KHC             & \$33.47 billion     & Consumer Staples       & Yes    \\           
\bottomrule
\end{tabular}
\caption{Summary of companies that appear in \textsc{FinanceBench}}
\label{tab:company_summary}
\end{table*}